\documentclass[10pt,comsoc]{IEEEtran} 
\usepackage{amsfonts}
\usepackage{times}
\usepackage{graphicx}
\usepackage{balance}  
\usepackage{graphicx}
\graphicspath{ {./figures} }
\usepackage{latexsym}
\usepackage{dsfont}
\usepackage{amssymb}
\usepackage{amsmath}
\usepackage{cite}
\usepackage{verbatim}
\usepackage{subfigure}

\usepackage{multirow}
\usepackage{amsmath}  
\usepackage{cite}
\usepackage{longtable}  
\usepackage{makecell}  

\def\bb0{{\mathbb{0}}}

\def\bb{{\mathbf{b}}}

\def\b0{{\mathbf{0}}}

\def\sf0{{\mathsf{0}}}

\usepackage{epstopdf}
\usepackage{enumerate}
\usepackage{algorithmicx}
\usepackage{algorithm}
\usepackage{amsmath}
\usepackage[noend]{algpseudocode}
\usepackage{float}
\usepackage{hyperref}
\usepackage{color}
\usepackage{makeidx}
\usepackage{bbm}
\usepackage{graphicx}
\usepackage{booktabs}
\usepackage{subfigure}
\usepackage{multirow}

\usepackage{stackengine} 
\stackMath

\newlength\matfield
\newlength\tmplength

\newcommand{\argmax}[1]{\underset{#1}{\text{argmax}}}

\usepackage{relsize}

\begin{document}
\title{Wireless Dataset Similarity: Measuring Distances in Supervised and Unsupervised Machine Learning}

\author{João Morais$^\dag$, Sadjad Alikhani$^\dag$, Akshay Malhotra$^\ddag$, Shahab Hamidi-Rad$^\ddag$, Ahmed Alkhateeb$^\dag$
 \\ 
\{joao, alikhani, alkhateeb\}@asu.edu , \{akshay.malhotra, shahab.hamidi-rad\}@interdigital.com }
\maketitle

\begin{abstract}
This paper introduces a task- and model-aware framework for measuring similarity between wireless datasets, enabling applications such as dataset selection/augmentation, simulation-to-real (sim2real) comparison, task-specific synthetic data generation, and informing decisions on model training/adaptation to new deployments. We evaluate candidate dataset distance metrics by how well they predict cross-dataset transferability: if two datasets have a small distance, a model trained on one should perform well on the other. We apply the framework on an unsupervised task, channel state information (CSI) compression, using autoencoders. Using metrics based on UMAP embeddings, combined with Wasserstein and Euclidean distances, we achieve Pearson correlations exceeding 0.85 between dataset distances and train-on-one/test-on-another task performance. We also apply the framework to a supervised beam prediction in the downlink using convolutional neural networks. For this task, we derive a label-aware distance by integrating supervised UMAP and penalties for dataset imbalance. Across both tasks, the resulting distances outperform traditional baselines and consistently exhibit stronger correlations with model transferability, supporting task-relevant comparisons between wireless datasets.
\end{abstract}

\vspace{-.15cm}

\section{Introduction}
Machine learning (ML) applications in wireless communications are seeing rapid growth \cite{8322184, 9370097, 8395149, 9121328, 8938771}. 
Seeking new frontiers in spectral efficiency, engineers and researchers have turned to the latest advances in computer sciences, especially those leveraging neural networks. 
The attention that learning mechanisms such as neural networks have received is evident in the several-fold growth in yearly submissions of AI/ML works to wireless conferences. 
Present research efforts, however, place excessive emphasis on the learning approach and arguably less so on the data used for learning. 
As a result, most machine learning research in wireless has yet to leave the academic setting and be deployed in real-world settings. 
To address this gap, some important questions regarding model design, training, and transition to deployment need to be addressed first:

\begin{itemize}
    \item How to choose adequate datasets for training models?
    \item How to predict model performance in real deployments?
    \item How to measure and ensure model generalization across different datasets?
\end{itemize}

Attempting to answer these questions, this work aims to provide an overview of \textit{dataset similarity} and its application to wireless communications. Dataset similarity/distancing \cite{inproceedings_datasetSim} consists of measuring how close (similar) or how far (different) two datasets are. In doing so, we aim to develop a method to estimate the generalization performance of an ML model from one dataset to another, without requiring model training.
Other potential uses of this research include i) detecting distribution shifts in data during real-world operation ii) improving transfer learning by selecting datasets with higher similarity (and thus transferability performance) to the target environments, and iii) enhancing real data augmentation with adequate synthetic data.
Indeed, obtaining real-world wireless data is challenging, making data augmentation essential for practical machine learning development and deployment. 

Data to train machine learning models can be obtained in a few different ways. The first is via real-world experimentation. Testbeds like POWDER \cite{powder} and AERPAW \cite{Guvenc2023AERPAW} provide experimentation platforms for researchers to access and prototype with live deployments. The second is via real-world datasets, such as those from DeepSense6G \cite{DeepSense} or NYU \cite{itwillwork}. These two approaches provide valuable insights but typically lack the scale and diversity needed for comprehensive machine learning model development. This has led to extensive use of simulated data, which can be divided into two main approaches: stochastic and deterministic. 

Stochastic models, such as those defined by 3GPP (e.g., CDL, TDL, UMa, UMi), are widely adopted and accessible through tools like MATLAB’s 5G Toolbox, and simulate channel conditions probabilistically. On the deterministic side, two options exist. The first, ray tracing methods, supported by tools like Wireless InSite \cite{wireless_insite} and the more recent SionnaRT \cite{hoydis2023sionnartdifferentiableray}, offer high-fidelity, site-specific simulations, particularly important for 6G research. The second, datasets like DeepMIMO \cite{DeepMIMO}, which combine ray tracing with parametrized channel generation, offer a hybrid approach that provides additional flexibility to deterministic and site-specific ray tracing. However, despite the availability of these tools, the wireless community still faces a lack of robust data operations — understanding and comparing datasets — hindering progress in developing large-scale generative models.

Purely real-world datasets, or purely synthetic datasets, may not be sufficient for comprehensive model development. To address this, supplementing realistic datasets with simulated data becomes crucial. This can be done through ray tracing, but recreating an environment can be a costly effort, and doing so for every setting is impractical. Similarly, configuring stochastic models to match specific environments is complex. One potential solution is data augmentation, where instead of creating entirely new datasets for each scenario, we leverage existing datasets that may not perfectly match the environment but are distributionally similar. By carefully selecting or augmenting datasets, we can reduce the need for generating new data from scratch. This highlights the importance of assessing the similarity between datasets and data distributions - to ensure that the supplemental data enhances model performance without introducing inconsistencies.

\begin{figure}[t]
    \centerline{\includegraphics[width=1\columnwidth]{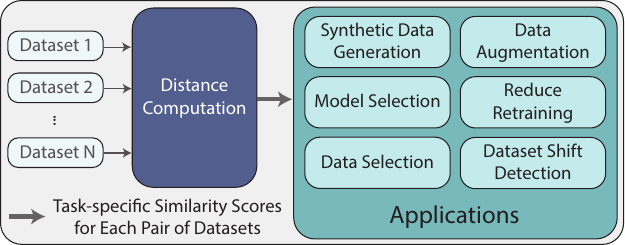}}
    \caption{Example applications enabled by dataset distance computation.}
    \label{fig:applications}
\end{figure}

\begin{figure*}[t]
	\centerline{\includegraphics[width=2\columnwidth]{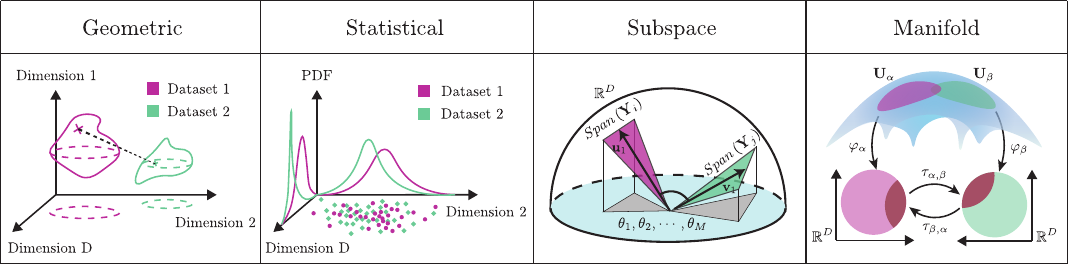}}
	\caption{Classes of distances that can be applied between datasets. }
	\label{fig:distances}
\end{figure*}

\textbf{Contribution:} To that end, in this work, we aim to provide wireless researchers means of comparing datasets before model training, and means of assessing whether model retraining is necessary, being purely synthetic or for real-world data augmentation. Figure \ref{fig:applications} shows how our dataset distance computation framework can be applied to a group of datasets to enable a wide range of applications. It can further help avoiding model retraining by predicting model performance from datasets alone. In summary, this work focuses on the following contributions: 
\begin{itemize}
    \item We develop a task-driven, model-agnostic framework that evaluates similarity between datasets, without the need for training additional models. 
    \item  We design two distance metrics that first apply a topology-preserving dimensionality reduction technique, like UMAP, to project data into a lower-dimensional space. Then, distances in the lower-dimensional space are computed: The first distance is Euclidean applies on clusters formed with KNN, while the second distance is based on Wasserstein and is applied between distributions of each dimension. Both approaches are evaluated on supervised and unsupervised tasks, showing clear improvements over previous methods.
    \item Specifically for supervised tasks, we propose a novel supervised distance computation method that effectively utilizes label information, introducing penalty terms to refine the accuracy of datapoint comparisons, increasing metric accuracy and robustness
    \item We demonstrate that our framework can be used to measure correlation between dataset distances computed with metrics and model performances, offering the potential create task-specific metrics to guide dataset choice, model retraining and benchmarking.
\end{itemize}

The implementation of the proposed framework is made open-source along with all evaluation scripts used in this work. Additionally, we provide thorough documentation and instructions to reproduce all research provided in this manuscript. \footnote{Documentation, artifacts, and reproducibility instructions can be found in the webpage: \textit{https://wi-lab.net/research/dataset-similarity}}

\textbf{Organization:} This work is organized as follows. Section \ref{sec:state} reviews the state of the art, exploring dataset similarity in machine learning and related fields, and discusses various types of distances and their desired properties. Section \ref{sec:problem} defines the problem, detailing what constitutes a dataset, the tasks, and the objectives of correlating distance metrics with model performance. In Section \ref{sec:unsup}, we apply our dataset distances to an unsupervised scenario, specifically CSI compression, to assess how these distances correlate with test performance. Section \ref{sec:sup} extends this application to supervised contexts, including LoS status and beam prediction. Finally, Section \ref{sec:conclusion} concludes with potential implications of this research and future work.

\section{State of the Art in Dataset Similarity Metrics} \label{sec:state}

Dataset similarity metrics are particularly relevant in the field of domain adaptation \cite{SIEGERT20181, ACHARA2024127205}. Domain adaptation is a subfield of machine learning that focuses on transferring knowledge from one domain (or dataset) to another, particularly when the two domains have different distributions \cite{NIPS2006_b1b0432c, 10.1007/978-3-030-71704-9_65, WANG2018135, Tzeng_2017_CVPR, 8892738}. It is widely applied when data from the target domain is scarce or difficult to obtain \cite{8861136}, but there is abundant data from a similar source domain. For example, in wireless communication, one might train a model on data collected in a specific urban environment but need to apply the same model to a rural or suburban setting, where the signal patterns and interference conditions differ \cite{265001, 9257198, 8487345}. Another case that may occur is when real-world data collections are infeasible, and models need to adapt from using data from simulations. Domain adaptation allows machine learning models to generalize better by learning from the similarities between these domains while adapting to their differences \cite{pinheiro2018unsuperviseddomainadaptationsimilarity, Fu_2019_ICCV, PENG2022462}. Central to this adaptation process is the use of dataset similarity metrics, which quantify how close or far apart two datasets are.

\textbf{Classes of Distances}: In many data-driven applications, choosing an appropriate distance metric is crucial for understanding the structure and relationships within the data. Distances can be categorized into several classes based on the nature of the data and the specific task at hand. Each category reflects different assumptions about how data points are organized, whether they lie on a linear subspace \cite{WANG2006456}, follow a distribution \cite{6832827}, or exist on a curved manifold \cite{umap_paper, tsne_paper, 10.1145/1273496.1273527}. Below, we explore four primary classes of distance metrics: geometric, statistical, subspace, and manifold-based, followed by a brief discussion of other types of distance metrics commonly used in machine learning and data analysis.

\textbf{Geometric Distances} measure direct spatial relationships between points in typically flat spaces like Euclidean space \cite{liberti2012euclideandistancegeometryapplications, DANIELSSON1980227, DEMAESSCHALCK20001}. These distances focus on the physical or spatial difference between data points, often assuming that the space is regular and linear. The most common geometric distance is \textit{Euclidean distance}, which calculates the straight-line distance between two points. Other examples include \textit{Manhattan distance}, which sums the absolute differences along each dimension, and \textit{cosine similarity}, which measures the angular distance between vectors. Geometric distances are intuitive and commonly used in tasks like clustering, regression, and classification when the relationships between data points do not involve curvature or probabilistic interpretations.

\textbf{Statistical Distances} quantify the difference between probability distributions or statistical properties of datasets \cite{6832827}. These distances are divided into divergences and integral probability metrics (IPMs). Divergences, such as \textit{Kullback-Leibler (KL) divergence} \cite{10.1214/aoms/1177729694, Narayana_1961}, measure how one distribution diverges from another, therefore tend to be asymmetric. IPMs,  like \textit{Wasserstein distance} \cite{Panaretos_2019} compute the minimal "cost" of transforming one distribution into another by moving probability mass. Other examples include \textit{Jensen-Shannon divergence}, a symmetric version of KL, and \textit{Maximum Mean Discrepancy (MMD)} \cite{dziugaite2015traininggenerativeneuralnetworks}, which compares distributions in a reproducing kernel Hilbert space. 

\textbf{Subspace Distances} are used to measure the relationships between subspaces rather than individual data points \cite{10.1145/1390156.1390204, ye2016schubertvarietiesdistancessubspaces}. In many high-dimensional tasks, data can be represented as subspaces, such as in signal processing or computer vision, where the key information lies in the orientation or span of the data rather than the individual points. \textit{Grassmannian distance} \cite{10.1145/1390156.1390204}, for example, measures the distance between subspaces on the Grassmann manifold, while \textit{principal angles distance} captures the angles between subspaces. Subspace distances are particularly useful in fields where data can be modeled as lying within a lower-dimensional subspace, such as MIMO systems in wireless communications, facial recognition and protein folding in bioinformatics. 

\textbf{Manifold-Based Distances} account for the fact that data often resides on a curved, nonlinear space rather than in a flat Euclidean space \cite{10.1145/1273496.1273527}. These distances aim to respect the intrinsic geometry of the data, capturing both local and global relationships. \textit{Geodesic distance}, for instance, measures the shortest path between two points along the manifold (often approximated via shortest paths on a neighborhood graph). \textit{Diffusion distance} reflects connectivity via random walks on such graphs. Relatedly, manifold learning methods such as \textit{t-SNE} \cite{tsne_paper} and \textit{UMAP} \cite{umap_paper} use neighborhood relationships to construct low-dimensional embeddings that preserve local structure; distances can then be computed in the learned representation space. 

\begin{table}[thbp]
\scriptsize  
\centering
\caption{Distance Calculation Approaches for Datasets $\mathbf{X}$ and $\mathbf{Y}$}
\begin{tabular}{|c|c|c|}
\hline
\textbf{Category} & \textbf{Metric} & \textbf{Distance Computation} \\ \hline

\multirow{4}{*}{\vspace{-0.8cm} Geometric} 
& \makecell{{Pair-wise}\\{Euclidean} \cite{euclidean_distance_2}} & 
\makecell{$\frac{1}{nm} \sum_{i=1}^{n} \sum_{j=1}^{m} \|\mathbf{X}_i - \mathbf{Y}_j\|_2$} \\ \cline{2-3}

& \makecell{{Centroid-wise}\\{Euclidean} \cite{e21020196}} & 
\makecell{$\lVert \bar{\mathbf{X}} - \bar{\mathbf{Y}} \rVert_2 $} \\ \cline{2-3}

& \makecell{{Cluster-wise}\\{Euclidean} \cite{electronics9081295}} & 
\makecell{$\frac{1}{c_1 c_2} \sum_{i=1}^{c_1} \sum_{j=1}^{c_2} \|\bar{\mathbf{X}}_{C,i} - \bar{\mathbf{Y}}_{C,j}\|_2$} \\ \cline{2-3}

& \makecell{Cosine Distance} \cite{XIA201539}& 
\makecell{$1 - \frac{\sum_{i=1}^{d} \mathbf{X}_i \mathbf{Y}_i}{\|\mathbf{X}\| \|\mathbf{Y}\|}$} \\ \hline

\multirow{8}{*}{\vspace{-2.4cm} Statistical} 
& \makecell{{Kullback-Leibler}\\{Divergence} \cite{kullback_leibler}} & 
\makecell{$\sum_i P(i) \log\left(\frac{P(i)}{Q(i)}\right)$} \\ \cline{2-3}

& \makecell{{Jensen-Shannon}\\{Divergence} \cite{1365067}} & 
\makecell{$\frac{1}{2} \sum_i P(i) \log\left(\frac{P(i)}{M(i)}\right)$ \\ $+ \frac{1}{2} \sum_i Q(i) \log\left(\frac{Q(i)}{M(i)}\right)$} \\ \cline{2-3}

& {Hellinger \cite{0fc2279d-14f3-3b9e-9bd3-45287989f2c1}} & 
\makecell{$\frac{1}{\sqrt{2}} \sqrt{\sum_i (\sqrt{P(i)} - \sqrt{Q(i)})^2}$} \\ \cline{2-3}

& \makecell{{Wasserstein}\\ \cite{Panaretos_2019}} & 
\makecell{$\inf_{\gamma \in \Gamma(P, Q)} \int |x - y| d\gamma(x, y)$} \\ \cline{2-3}

& \makecell{MMD} \cite{dziugaite2015traininggenerativeneuralnetworks} & 
\makecell{$\lVert \mathbb{E}_{X \sim P}[\phi(X)]$ \\ $- \mathbb{E}_{Y \sim Q}[\phi(Y)] \rVert^2$} \\ \cline{2-3}

& \makecell{{Kolmogorov} \\ -Smirnov} \cite{doi:10.1080/01621459.1951.10500769} & 
\makecell{$\sup_x |F_{\mathbf{X}}(x) - F_{\mathbf{Y}}(x)|$} \\ \cline{2-3}

& \makecell{Energy Distance} \cite{NEURIPS2020_9873eaad} & 
\makecell{$2 \mathbb{E}\lVert \mathbf{X} - \mathbf{Y} \rVert - \mathbb{E}\lVert \mathbf{X} - \mathbf{X}' \rVert $ \\ $- \mathbb{E}\lVert \mathbf{Y} - \mathbf{Y}' \rVert$} \\ \cline{2-3}

& \makecell{{Total Variation} \\ \cite{6804281}} & 
\makecell{$\frac{1}{2} \int |P(x) - Q(x)| dx$} \\ \hline

\multirow{4}{*}{Subspace} 
& {Grassmann \cite{Bendokat_2024}} & \makecell{$\lVert \mathbf{\theta} \rVert_2 $} \\ \cline{2-3}

& {Chordal \cite{mankovich2023chordalaveragingflagmanifolds}} & \makecell{$\sqrt{\sum_{i=1}^{k} \sin^2(\theta_i)}$}  \\ \cline{2-3}  

& {Asimov \cite{9024000}} & \makecell{$\theta_k$} \\ \hline 

\iftrue
\multirow{3}{*}{Manifold} 
& {PCA} \cite{MACKIEWICZ1993303} & \makecell{Embedding: Linear Projection \\ (min. reconstruction error)} \\ \cline{2-3}

& {t-SNE \cite{tsne_paper}} & \makecell{Embedding: Preserve Neighborhoods \\ (min. Kullback-Leibler Divergence)} \\ \cline{2-3}

& {UMAP \cite{umap_paper}} & \makecell{Embedding: Preserve Topology \\ (neighbor-graph; min.  cross-entropy)} \\ \hline
\fi

\end{tabular}
\end{table}

Some distances are used in specialized applications, such as the \textit{Proxy A-Distance (PAD)} \cite{LostDomainGeneralization}, which measures the distinguishability between two datasets based on the classification error of a model trained to differentiate them. A lower error indicates a smaller distance, making PAD useful in domain adaptation. Another important metric is the \textit{Maximum Mean Discrepancy (MMD)} \cite{dziugaite2015traininggenerativeneuralnetworks}, a kernel-based distance used to compare distributions by analyzing their means in a reproducing kernel Hilbert space \cite{alvarez2012kernelsvectorvaluedfunctionsreview}. MMD is frequently employed in generative modeling and distribution alignment tasks. 

Figure \ref{fig:distances} summarizes the concepts behind various classes of distances. However, understanding these concepts or formulas does not fully explain their effectiveness in computing dataset distances, especially when these distances are expected to correlate with model performance. In the following sections, we will mathematically define our problem and describe the evaluation of distances. We will examine which properties of distances are generally desirable and which properties of the dataset may help choose a given distance. 

\section{Calculation of Dataset Distances in Practice} \label{sec:practice}
This section defines mathematically the fundamental terms needed to explore dataset similarities/distances. These terms include a dataset, a distance, and several clear ways to apply distances to datasets, namely considering datasets as matrices, applying pairwise functions, clustering points before applying such functions, and estimating per-feature or joint data distributions and comparing such distributions. Additionally, this section shows how distance computation would work in a latent space after applying a dimensionality reduction method to the original dataset, and why this can be beneficial. Let us start with a general definition of a dataset.

\textbf{Dataset:} A dataset consists of a collection of \( N \)-dimensional datapoints, where each datapoint is represented as a vector \( \mathbf{x} = [x_1, x_2, \dots, x_N] \in \mathbb{R}^N \). This vectorized form is general and flexible, allowing any type of data to be reshaped into this format. Complex-valued data can be converted by separating real and imaginary parts. Tensors of higher dimensions, like images, can be flattened. A dataset $\mathcal{D}$ with $M$ datapoints can be written as a set 

\[
\mathcal{D} = \{\mathbf{x}_j\}_{j=1}^{M}, \quad \mathbf{x}_j \in \mathbb{R}^N,
\]
or be represented as a $M \times N$ matrix, where each row corresponds to a datapoint and each column to a feature. 

\textbf{Dataset Distance:} A dataset distance \( d \) can be defined as a function that operates between two datasets and outputs a non-negative number, i.e., 

\[
d(\mathcal{D}_1, \mathcal{D}_2): \mathbb{R}^{M_1 \times N} \times \mathbb{R}^{M_2 \times N} \rightarrow \mathbb{R}^1. 
\]

Here, \( M_1 \) and \( M_2 \) represent the number of datapoints in datasets \( \mathcal{D}_1 \) and \( \mathcal{D}_2 \), respectively. Note the requirement that both datasets need to have the same number of features, \( N \).
 
\textbf{Matrix Distances:} Based on this definition, the first and most straightforward way to compute distances between datasets is to apply matrix distances. A common approach is to compute the difference between two matrices and apply a matrix norm, such as the Frobenius norm, to the result. However, this method has two limitations: i) it requires the datasets to have the same number of datapoints, and ii) shuffling the datapoints within each dataset can lead to different distance values. 

\textbf{Pairwise Distances}: The second way is to compute a distance between each pair of points and accumulate or average it across all pairs. This is called a pairwise distance (typically geometric) and can be written as 

\begin{equation}
    d_p(\mathcal{D}_1, \mathcal{D}_2) = \sum_{j}^{M_1} \sum_{k}^{M_2} f\left(\mathbf{x}_j^{(1)},\mathbf{x}_k^{(2)}\right)
\end{equation}
where $\mathbf{x}_j^{(i)}$ represents the $j$-th datapoint from the dataset $\mathcal{D}_i$. The function $f: \mathbb{R}^N \times \mathbb{R}^N \rightarrow \mathbb{R}^1$ takes a two $N$-dimensional point from the dataset and returns a single non-negative numeral. This approach has the challenge of requiring a very accurate weighting between points, especially in high dimensions. The function $f$ can further be applied over clusters of data, instead of single datapoints. The drawback is requiring a clustering technique, which inherently requires assumptions on distance metrics to create these clusters, and will introduce the need for new computations. Nonetheless, clustering provide averaging, leading to more robustness in high dimensions and larger datasets - useful features when applying the distance metric to datasets.

\textbf{Distribution Distances}: The last alternative for computing distances between datasets is to consider joint or per-feature distributions (typically with statistical distances). In the simplest case, the per-feature empirical distributions take the shape of histogram, which can be represented as a vector where each entry corresponds to the frequency of datapoints in the bin. As noted in \cite{CHA20021355_Distances_between_histograms}, the distance function 

\begin{equation}
    d_H(\mathcal{D}_1, \mathcal{D}_2) = d(H(\mathcal{D}_1), H(\mathcal{D}_2)),    
\end{equation}
where the histogram $H$ assumes parameters such as bin size, which may vary across features. The choice can be made heuristically - the histogram bins should be as small as possible to offer enough resolution when comparing distributions, but sufficiently large to contain enough samples and be robust to outliers. As such, the creation of the histogram is done behind the scenes, as is often done in most implementations. Here we also assume the distribution is captured accurately. Note, however, that the accuracy of this distance can decay significantly for small, high-variance datasets. Nonetheless, several distances are successful comparing distributions, like the Energy and Wasserstein / Earth's Mover Distance (EMD).  

\textbf{Latent Space Projections:} Transforming datasets into a lower-dimensional latent space, \( T(\mathcal{D}) \), offers two key advantages: it improves the accuracy of distance calculations by emphasizing the most important features while eliminating redundant ones, and it makes these calculations more computationally tractable. For example, a dataset represented by a \(32 \times 32\) channel matrix, when flattened, produces a \(1024\)-dimensional vector. Calculating distances in such high-dimensional spaces can be inaccurate since the dataset needs to be very large for a proper distribution estimation. By applying a transformation \( T \), the data is projected into a lower-dimensional space where the distance function 

\begin{equation}
d_T(\mathcal{D}_1, \mathcal{D}_2) = d(T(\mathcal{D}_1), T(\mathcal{D}_2))
\end{equation}
is computationally efficient and reflective of meaningful data relationships.

Lowering the dimensionality is particularly beneficial for distribution-based distances, such as the Wasserstein distance, which rely on estimating multi-dimensional distributions or histograms. In high dimensions, accurately estimating these distributions requires significantly more data. Reducing dimensionality simplifies the estimation process, making it more feasible to compute accurate distances between datasets. Thus, latent space projections not only reduce the computational burden but also ensure that distances better capture the intrinsic properties of the data, enhancing interpretability and improving performance predictions.

Now that the tools for operating on datasets are defined, we can describe the challenge these distances attempt to solve.

\section{Dataset Similarity Problem Definition} \label{sec:problem}

In this section, we define the dataset similarity problem: to compute distances between datasets that correlate strongly with model performance across those datasets. Thus if two datasets are similar, the dataset distance should be small, and the ML model trained on one dataset should generalize well to the other. Whereas in the distance is large, the generalization performance is expected to be poor. Solving this would enable us to predict how well a model trained on one dataset performs on another, without exhaustive training and testing. As mentioned, this valuable for transfer learning, domain adaptation, and model selection or switching. To frame the problem, we how machine learning models are trained and tested on dataset and how correlation between test scores is computed when using many datasets - this computation is required measure the quality of a distance metric. The end goal is to create a distance metric that accurately reflects dataset relationships in terms of task performance.



\begin{figure*} [t]
\centerline{\includegraphics[width=2\columnwidth]{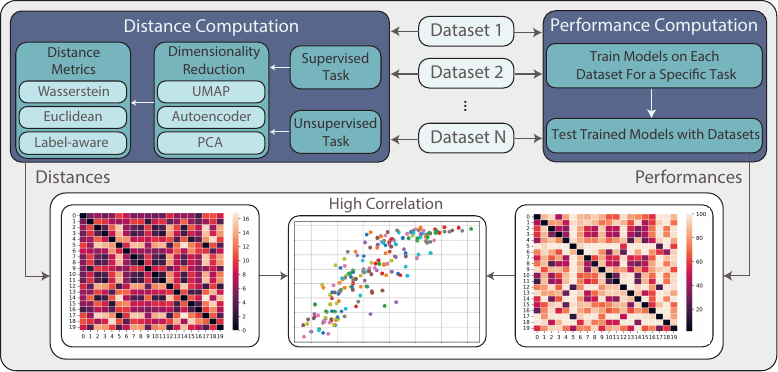}}
\caption{Framework for evaluating a dataset distance metric for a given task and model: if two datasets are close according to the metric, then a model trained on one should achieve high performance when tested on the other. The higher the correlation between distances computed with a given metric and model performance on a given task, the more suitable the metric is for measuring dataset similarity in that task.}
\label{fig:system_model}
\end{figure*} 

\textbf{Task-specific Model Training:} The model definition (architecture), choice of loss function, training parameters, data splits, and other configurations, all depend on the particular task and dataset at hand. for simplicity, we represent only the dependencies with task and dataset. First, we define a machine learning model $\mathcal{M}$ built towards a task $\mathcal{T}$ and trained on a dataset $\mathcal{D}$ as 

\begin{equation} \label{eq:task_train}
    \mathcal{M}^\mathcal{T}_\mathcal{D} = \mathcal{M}^\mathcal{T \, \text{train}} \left(\mathcal{D}\right).
\end{equation}
Further note that the training operation is performed often on a subset of features from the dataset $\mathcal{D}$, called the input features. In case of unsupervised tasks, all features may be input features. In supervised tasks, the features used for model inputs and outputs are different.

\textbf{Task-specific Model Inference:} After training successfully, i.e., by seeing an adequate decrease of training loss, then providing inputs similar to the training dataset to the model is expected to result in outputs that resemble the respective output training data. The next step is inferencing this model on possibly different datasets. This step is relevant to measure model transferability. Supposing a model is trained on source dataset $\mathcal{D}_S$ and tested on target $\mathcal{D}_T$, the model outputs can be given by

\begin{equation} \label{eq:task_test}
    \mathcal{M}^\mathcal{T}_{\mathcal{D}_S \rightarrow \mathcal{D}_T} = \mathcal{M}^{\mathcal{T}  \, \text{test}}_ {\mathcal{D}_S} \left(\mathcal{D}_T\right).
\end{equation}
These outputs can subsequently be used in loss functions to determine model performance on the task. 

\textbf{Transferability Performance Metric:} The loss $L$ quantifies how well the model performed on the target dataset. In this context, $L$ can be used to find which source dataset $\mathcal{D}_S$ is more suitable to train $\mathcal{M}_\mathcal{T}$ for inference in the target dataset $\mathcal{D}_T$. The choice is made by selecting the source dataset that lead to the highest performance in the target dataset. Or, equivalently, the dataset that resulted in the lowest performance drop compared to the ideal performance, i.e., when the model is trained in the target dataset. Mathematically, we can define the model transferability performance $\mathcal{P}$ between two datasets by

\begin{equation}
    \mathcal{P}(\mathcal{D}_i, \mathcal{D}_j) = L\left(\mathcal{M}^\mathcal{T}_{\mathcal{D}_j \rightarrow \mathcal{D}_j}\right) - L\left(\mathcal{M}^\mathcal{T}_{\mathcal{D}_i \rightarrow \mathcal{D}_j}\right)
\end{equation}

\textbf{Distances and performance matrices:} To correlate distances and performances, we require several examples of each, for which is necessary multiple datasets. Considering $K$ datasets, we may aggregate distances and transferability performances in matrices: 

\begin{gather} \label{eq:fim2}
	\mathbf{D} = 
	\begin{bmatrix}
		d_{i,j}
	\end{bmatrix} \ : \ d_{i,j} = d(\mathcal{D}_i, \mathcal{D}_j)\\ 
	\mathbf{P} = 
	\begin{bmatrix}
		p_{i,j}
	\end{bmatrix} \ : \ p_{i,j} = \mathcal{P}(\mathcal{D}_i, \mathcal{D}_j)
\end{gather}
with $i,j = 1, \dots, K$. These matrices can be plotted as confusion matrices, which done in Section \ref{sec:unsup} in Figure \ref{fig:system_model}. The formulation presented here permits taking the elements of each matrix for correlation calculations.

\textbf{Correlation Between Distances and Performances:} Let \( \rho(\mathbf{d}, \mathbf{p}) \) denote the correlation coefficient between the distance vector \( \mathbf{d} \) and the performance vector \( \mathbf{p} \), which respectively represent the distances and model performances for pairs of datasets. We define $\textbf{d}$ and $\textbf{p}$, as
\begin{gather}
	\textbf{d} = \text{vec}(\mathbf{D}) = \begin{bmatrix} d_{1,1}, ..., d_{1,K}, d_{2,1}, ..., d_{2,K}, ..., d_{K,K}  \end{bmatrix} \\
	\textbf{p} = \text{vec}(\mathbf{P}) = \begin{bmatrix} p_{1,1}, ..., p_{1,K}, p_{2,1}, ..., p_{2,K}, ..., p_{K,K}  \end{bmatrix},
\end{gather}
making our objective possible to be written as 

\begin{equation} \label{eq:obj}
    d_\mathcal{T}^\star = \argmax{d_\mathcal{T}} \ \rho(\mathbf{d}, \mathbf{p})
\end{equation}
where the distances vector $\mathbf{d}$ depends of the distance function $d_\mathcal{T}$ and the datasets $\boldsymbol{\mathcal{D}} = \{\mathcal{D}_1, ..., \mathcal{D}_K\}$. The performances vector $\mathbf{p}$ is a function of $\boldsymbol{\mathcal{D}}$ and the model $\mathcal{M}$. The problem consists in finding an optimal distances $d_\mathcal{T}^\star$ that maximizes the correlation between the distances distances produced by that function and the performances. 

Parting from equation \eqref{eq:obj}, next sections present the analysis framework for computing dataset distances and comparing them with model performances. Then, subsequent sections take this framework and apply it to specific datasets and tasks to assess whether the correlation between the computed distances and performances is high. Our goal is to derive a general method for creating effective, task-specific and low-complexity dataset distances. 

\section{Framework for Evaluating Dataset Distances and Model Performance Correlation} \label{sec:framework}

To address the challenge of selecting distance functions that accurately predict model performance across different datasets, we propose a comprehensive framework that correlates computed dataset distances with model performance metrics. This framework evaluates the suitability of various distance measures by analyzing their ability to reflect performance degradation when transferring models between datasets. By effectively measuring dataset similarities that reflect on model discrepancies, our framework can be used to develop distances for detecting dataset shifts, guiding generative data augmentation, and ranking the most useful datasets for a particular task. 

\textbf{Framework overview}: In this framework, each dataset undergoes two primary processes: distance computation and performance evaluation. The distance computation step calculates a distance metric \( d \) between pairs of datasets, resulting in a distance matrix \( \mathbf{D} \). Performance evaluation involves training a machine learning model on one dataset and testing it on others, yielding a performance matrix \( \mathbf{P} \). By vectorizing and correlating \( \mathbf{D} \) and \( \mathbf{P} \), we can identify patterns and assess how well the distances predict model performance drops. A schematic representation of this framework is provided in Figure~\ref{fig:system_model}.

\textbf{Performance Computation}: The performance evaluation begins by training a model \( M_i \) on each dataset \( D_i \) and assessing its performance \( P_{ii} \) on a test set from the same dataset. This establishes a baseline performance for the model within its training domain. The trained model \( M_i \) is then tested on other datasets \( D_j \) (for \( j \neq i \)), and the performance \( P_{ij} \) is recorded. If two datasets \( D_i \) and \( D_j \) are similar, we expect \( P_{ij} \) to be close to \( P_{ii} \); conversely, a significant drop in performance indicates substantial differences between the datasets. The objective is to evaluate whether large distances \( d_{ij} \) between datasets correspond to large performance drops \( \Delta P_{ij} = P_{ii} - P_{ij} \).

\textbf{Distance Computation:} The distance computation step calculates a scalar distance \( d_{ij} \) between each pair of datasets \( D_i \) and \( D_j \). Various distance measures can be employed, including statistical distances based on empirical probability density functions (PDFs) or distances computed in transformed spaces (e.g., after dimensionality reduction). A critical aspect of computing statistical distances such as Jensen-Shannon divergence or Maximum Mean Discrepancy (MMD) is the accurate estimation of the empirical PDFs for the datasets.

\textbf{Jointly estimating dataset PDFs}: The main challenge lies in estimating the empirical PDFs jointly between both datasets to ensure comparability across datasets. This requires constructing histograms with identical bin edges for both datasets. Let \( D_i = \{ \mathbf{x}_n^{(i)} \}_{n=1}^{M_i} \) and \( D_j = \{ \mathbf{x}_n^{(j)} \}_{n=1}^{M_j} \) be the data samples from datasets \( D_i \) and \( D_j \), respectively, where \( M_i \) and \( M_j \) are the number of samples in each dataset.

\begin{figure*} [t]
	\centerline{\includegraphics[width=2\columnwidth]{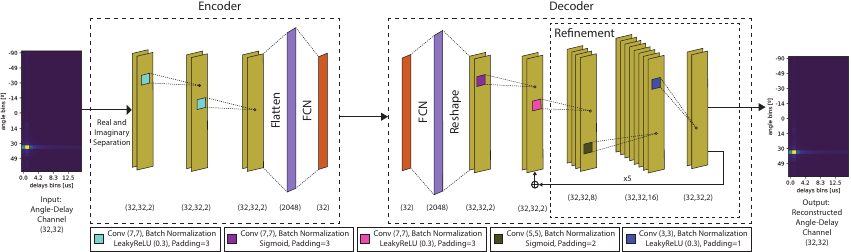}}
	\caption{Architecture of the model used in the unsupervised CSI compression task. The model is heavily inspired in the CSINet+ \cite{8972904}.} 
	\label{fig:AE}
\end{figure*}

We define a common set of bin edges \( \{ b_k \}_{k=0}^{K} \) that span the combined range of \( D_i \) and \( D_j \):
\begin{gather}
    b_0 = \min\left(\min(D_i), \min(D_j)\right), \\ 
    b_K = \max\left(\max(D_i), \max(D_j)\right),    
\end{gather}
with \( K \) being the number of bins, which can be set using a heuristic such as \( K = \sqrt{M} \), where \( M = \max(M_i, M_j) \). The empirical PDFs are then computed as normalized histograms per dataset \( D_i \):

\begin{equation}
h_i(k) = \sum_{n=1}^{M_i} \delta\left(b_{k-1} \leq \mathbf{x}_n^{(i)} < b_k\right), \quad k = 1, 2, \dots, K,
\end{equation}
where \( \delta(\cdot) \) is the indicator function. Subsequently, we normalize histogram counts to obtain the empirical PDF. 

\begin{gather}
p_i(k) = \frac{h_i(k)}{\sum_{k=1}^{K} h_i(k)} = \frac{h_i(k)}{M_i}, \\
p_j(k) = \frac{h_j(k)}{\sum_{k=1}^{K} h_j(k)} = \frac{h_j(k)}{M_j}.
\end{gather}
By using common bin edges and the same number of bins, we ensure that \( p_i \) and \( p_j \) are directly comparable. The choice of \( K \) and \( \{ b_k \} \) is critical: too many bins may result in empty bins and poor PDF estimation, while too few bins may oversmooth the distributions, obscuring important differences.

\textbf{Flexibility and Challenges}: Our framework is flexible, allowing for preprocessing steps such as dimensionality reduction or clustering, depending on the task and dataset characteristics. High-dimensional datasets pose challenges for distance computation due to the curse of dimensionality. Dimensionality reduction techniques (e.g., PCA, UMAP) can alleviate these challenges by projecting data into lower-dimensional spaces where distances are more meaningful and computationally tractable. Furthermore, clustering can further simplify the problem by summarizing datasets through cluster centroids, reducing computational complexity. However, it's important to consider that some clustering algorithms may introduce additional computational overhead. 

\textbf{Dependencies and Considerations:} The effectiveness of the framework depends on several factors:

\begin{itemize}
    \item \textbf{Dataset Characteristics}: Variations in data distributions, sample sizes, and feature spaces can impact distance computations and model performance.
    \item \textbf{Task Nature}: Supervised and unsupervised tasks may require different approaches for distance computation (e.g., label-aware vs. label-agnostic methods).
    \item \textbf{Model Selection}: Using consistent models across datasets minimizes model-dependent variability, allowing for a clearer analysis of dataset similarities.
\end{itemize}

By carefully considering these dependencies, we aim to ensure that the computed distances accurately reflect model performance differences. In the subsequent sections, we will apply this framework to both supervised and unsupervised tasks, evaluating various distance measures using the empirical PDF estimation methods described. By thoroughly assessing potential candidates for dataset distancing, we aim to identify the most effective distance metrics and generalize our findings to other domains.

\section{Dataset Distancing through Latent Space Projections}

In high-dimensional datasets, computing distances that accurately reflect the relationships between datasets and correlate with model performance is a challenging task, largely due to the presence of noise, irrelevant features, and the complexity of the data. Traditional distance metrics, when applied in their native high-dimensional space, frequently fail to capture the crucial underlying structures due to the presence of noise and irrelevant features. To address this, it is essential to map the datasets into a transformed space that emphasizes the most relevant features for the task at hand, akin to how dimensionality reduction techniques like the Johnson-Lindenstrauss (JL) \cite{Dasgupta2003AnEP} theorem focus on preserving distances, or how PCA \cite{MACKIEWICZ1993303} aims to increase discriminability. The transformation we need should preserve \textbf{local proximity}—ensuring that datasets close in terms of task-relevant features stay close in the new space—and also retain the \textbf{global structure}, allowing the broader relationships between datasets to be maintained. By doing so, we can compute distances that are more meaningful for the task, leading to higher correlations with model performance across different datasets.

This transformation can be achieved through a \textbf{graph-based approach}, where the relationships between datasets are modeled based on their local neighborhoods and global connections. The graph captures how datasets are related in terms of task-relevant features, ensuring that those with similar characteristics are clustered together in the transformed space. At the same time, this method maintains global structure, ensuring that datasets that are further apart, yet share broader similarities important to the task, are also represented appropriately. As a result, the transformed space becomes a \textbf{manifold-like representation} where distances between datasets reflect not only their proximity in feature space but also their structural properties, depending on the task. This enables us to compute distances that are more likely to correlate with how models trained on one dataset will perform on another, ultimately improving transfer learning, domain adaptation, and model selection. This task-aware approach to transforming the data ensures that we retain the most meaningful information while filtering out noise, leading to more accurate and efficient comparisons of datasets.

In this section, we introduce a novel method for measuring distances between datasets by leveraging \textbf{Uniform Manifold Approximation and Projection (UMAP)} to transform the original data into an encoded latent space. As previously defined earlier, let us consider datasets \( D_i = \{ \mathbf{x}_j^{(i)} \}_{j=1}^{M_i} \) with $M_i$ datapoints each dataset with \( \mathbf{x}_j^{(i)} \in \mathbb{R}^N \). Our objective is to compute the distance between these datasets in a space that captures their intrinsic geometric and topological properties more effectively than the raw feature space.

\textbf{Use of UMAP for latent spaces}: We employ UMAP to project the high-dimensional data into a lower-dimensional latent space. This transformation is defined by a function \( f_{\text{UMAP}}: \mathbb{R}^N \rightarrow \mathbb{R}^d \), where \( d \ll N \). The encoded datasets are then represented as \( \tilde{D}_i = \{ \tilde{\mathbf{x}}_j^{(i)} = f_{\text{UMAP}}(\mathbf{x}_j^{(i)}) \}_{j=1}^{M_i} \). UMAP constructs a fuzzy topological representation by building a weighted k-nearest neighbor graph in the high-dimensional space, capturing both local and global data structure. Fuzzy representations (i.e. non-binary set ownership relations) are leveraged in part because they allow a smoother cost function for iterative optimization. Using weighted set ownership relations of fuzzy sets, UMAP then optimizes a low-dimensional embedding that preserves this fuzzy simplicial set. The steps in the UMAP algorithm that enable this include:

\begin{enumerate}
    \item \textbf{Constructing a fuzzy simplicial set from high-dimensional data}: UMAP builds a weighted graph where the weights represent the probabilities of connection between data points. This effectively captures the local neighborhood relationships.
    \item \textbf{Defining a fuzzy topological representation}: Then, by considering these probabilities, UMAP creates a fuzzy topological space that represents the broader data manifold structure.
    \item \textbf{Optimization in Low-Dimensional Space}: UMAP finds a low-dimensional embedding that minimizes the cross-entropy between the fuzzy topological representations in the high-dimensional and low-dimensional spaces.
\end{enumerate}

By using this approach, UMAP preserves more of the global and local structure of the data compared to other manifold learning techniques. The reasons for this are: i) PCA is linear and may not capture non-linear structures; ii) t-SNE focuses on local neighborhoods and may distort global structures, and iii) autoencoders depend heavily on the network architecture and training process, focusing more in data reconstruction than inner relationship maintenance.

\textbf{Euclidean in UMAP spaces:} In the latent space, we explore different flavors of the Euclidean distance to quantify the separation between \( D_1 \) and \( D_2 \):

\begin{enumerate}
    \item \textbf{Pairwise Euclidean Distance}: Calculated between all pairs of points from the two datasets,
    \begin{equation}
    d_{\text{pairwise}} = \frac{1}{M_1 M_2} \sum_{j=1}^{M_1} \sum_{k=1}^{M_2} \left\| \tilde{\mathbf{x}}_j^{(1)} - \tilde{\mathbf{x}}_k^{(2)} \right\|_2.
    \end{equation}
    This metric provides a comprehensive measure by considering all possible point-to-point distances.
    
    \item \textbf{Cluster-Based Euclidean Distance}: We cluster each encoded dataset \( \tilde{D}_i \) into \( K_i \) clusters using a clustering algorithm such as k-means \cite{IKOTUN2023178}, hierarchical clustering \cite{cohenaddad2017hierarchicalclusteringobjectivefunctions}, or DBSCAN \cite{10.5555/3001460.3001507}. Each cluster is represented by its centroid \( \mathbf{c}_l^{(i)} \), calculated as the average of the points in that cluster:
    \begin{equation}
    \mathbf{c}_l^{(i)} = \frac{1}{|C_l^{(i)}|} \sum_{\tilde{\mathbf{x}} \in C_l^{(i)}} \tilde{\mathbf{x}},
    \end{equation}
    where \( C_l^{(i)} \) is the set of points in cluster \( l \) of dataset \( D_i \), and \( |C_l^{(i)}| \) is the number of points in that cluster. The cluster-based Euclidean distance is then defined as:
    \begin{equation}
    d_{\text{cluster}} = \frac{1}{K_1 K_2} \sum_{l=1}^{K_1} \sum_{m=1}^{K_2} \left\| \mathbf{c}_l^{(1)} - \mathbf{c}_m^{(2)} \right\|_2.
    \end{equation}
    This approach highlights structural differences at a cluster level, but its computation complexity is dependent on the clustering technique employed.
    
    \item \textbf{Average Euclidean Distance}: A special case of cluster-based distance when \( K_1 = K_2 = 1 \), simplifying to the distance between the mean vectors of the datasets,
    \begin{equation}
    d_{\text{average}} = \left\| \bar{\mathbf{x}}^{(1)} - \bar{\mathbf{x}}^{(2)} \right\|_2, \hspace{0.5em} \text{where} \hspace{0.5em} \bar{\mathbf{x}}^{(i)} = \frac{1}{M_i} \sum_{j=1}^{M_i} \tilde{\mathbf{x}}_j^{(i)}.
    \end{equation}
    This provides a coarse but the most computationally efficient measure of dataset separation.
\end{enumerate}

\textbf{Wasserstein in UMAP spaces:} Beyond Euclidean metrics, we compute the Wasserstein (Earth Mover's) distance between the datasets in the latent space by treating each dataset as a multivariate distribution. To simplify the computation, we decouple these distributions on a per-dimension basis, allowing us to compute one-dimensional Wasserstein distances for each dimension \( n \) in the latent space. The one-dimensional Wasserstein distance between the empirical distributions of the \( n \)-th dimension of datasets \( D_1 \) and \( D_2 \) is defined as:
\begin{equation}
W_n = \int_{0}^{1} \left| F_n^{(1)^{-1}}(t) - F_n^{(2)^{-1}}(t) \right| dt,
\end{equation}
where \( F_n^{(i)^{-1}}(t) \) is the inverse cumulative distribution function (quantile function) of the \( n \)-th dimension of dataset \( D_i \). Intuitively, the Wasserstein distance measures the minimal cost of transporting the mass of one distribution to match another, where cost is quantified by the amount of probability mass (earth)  moved times the distance it is moved.

To clarify, the integral computes the area between the two cumulative distribution functions (CDFs) of the datasets along dimension \( n \). This represents the average distance that a unit of probability mass must be moved to transform one distribution into the other. The overall Wasserstein distance between the datasets is then the average over all dimensions:
\begin{equation}
W = \frac{1}{d} \sum_{n=1}^{d} W_n.
\end{equation}
This metric effectively captures the distributional differences between the datasets across all dimensions, providing a robust measure of their similarity. It outshines the Euclidean distances when i) the datasets have enough points for an accurate estimation of empirical distributions; and ii) when the number of dimensions is low, since many dimensions can dilute the distribution distance value computed in the more relevant features. We see next how to further augment this distance method by making it robust to high-dimensional latent spaces.

\textbf{Augmenting with Feature Importances}: It is possible to expand both the Euclidean and Wasserstein metrics when some features in the latent space are known to be more important than others. Feature importance can be assessed using methods such as permutation feature importance (PFI) \cite{article_permutation}, partial dependence (PD) \cite{Molnar_2023} plots, Local Interpretable Model-Agnostic Explanations (LIME) \cite{ribeiro2016whyitrustyou}, or SHAP (SHapley Additive exPlanations) \cite{lundberg2017unifiedapproachinterpretingmodel} values. By weighting the distances according to feature importance, we obtain a more nuanced measure that reflects the relative significance of each dimension.

\textbf{Important considerations}: We utilize UMAP to create a latent space that faithfully represents the underlying structure of datasets. However, this application of UMAP requires attention to some details. First, UMAP can be computationally expensive when applied to high dimensional input spaces. For this reason, it can pay off to first reduce the input space using PCA or other computationally cheap (likely linear) dimensionality reduction techniques, and then using UMAP on this intermediate latent space. Additionally, UMAP requires a distance function to compare datapoints. For wireless channels, we found that the best distance function is a correlation distance. However, a more efficient euclidean distance can be used too if the complexity and scalability are more important than performance. Moreover, note that when applying UMAP to a set of datsets, all datasets should be processed jointly in a single UMAP fuzzy sets. Otherwise there is no guarantee the independent encodings will be consistent with one another, leading to likely wrong distances. 

\textbf{UMAP parameters}: It is key to set appropriate numbers for parameters like the minimum distance in the latent space and the number of neighbors. The number of neighbors in a KNN context determines how many of the closests neighbors will influence each computation in the UMAP. Larger number of neighbors lead to higher prioritization of global structures. Less neighbors results in better local structures at a cost of global understanding. A balance is key and we found that $32$ neighbors worked well for datasets considered, each having roughly $5000$ samples. The minimum distance between points should be kept low to allow the broader data relations to shine through. In other words, smaller distances allow datapoints to be represented closer together if they are similar in their relationship. We opt of a value of $0.1$ for the minimum distance in the latent space.

Overall, this approach offers significant advantages over traditional methods, providing a more detailed and interpretable assessment of dataset similarity. Next we apply the proposed distances and compare them with traditional methods in a unsupervised CSI compression task.

\section{Unsupervised case: CSI Compression} \label{sec:unsup}

This section delves into the application of distances to datasets in the context of one unsupervised machine learning task, CSI compression. We choose to begin with an unsupervised learning task because the data used for model input and as labels is the same, i.e. $\mathcal{N}_\mathcal{T}^I = \mathcal{N}_\mathcal{T}^O = \mathcal{N}$. In this task, we explore the correlation between distances and model performances. In summary, the variables defined in Section \ref{sec:problem} take the following values:

\begin{itemize}
	\item $\mathcal{T}$: CSI compression
	\item $\mathcal{P}$: NMSE between encoded channel and true channel
	\item $\mathcal{M}$: Convolutional autoencoder, architecture in Figure \ref{fig:AE}
	\item $\boldsymbol{\mathcal{D}}$: $20$ areas from the ASU Campus DeepMIMO dataset
\end{itemize}

In the remaining of the section first describes the variables in more detail, maintaining the order showed above. After clarifying the conditions in which the framework of Section \ref{sec:framework} is applied, we show the results obtained and discuss.

\begin{figure*} [t]
	\centerline{\includegraphics[width=2\columnwidth]{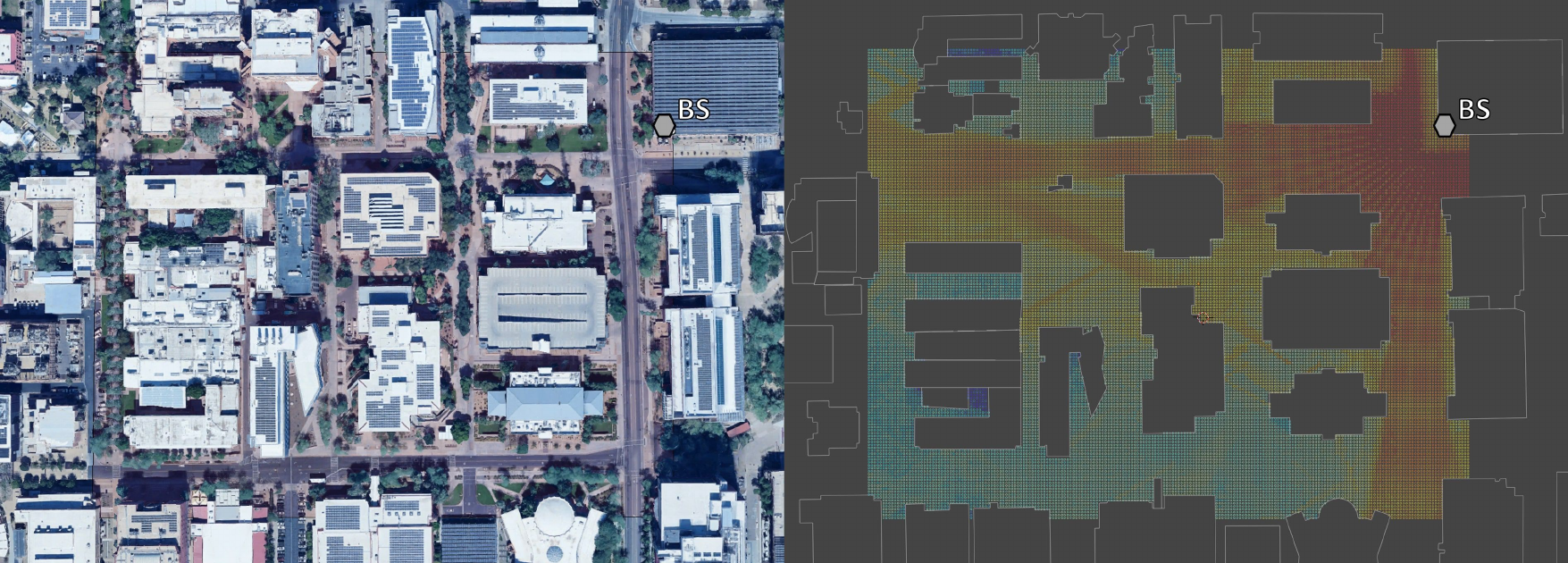}} 
	\caption{Real (left) and rendered (right) top view the ASU campus DeepMIMO dataset. The base station is showed in both figures. It should be noted that buildings and other scenario assets are 3D, and their heights matter significantly for roof diffractions. The mesh represented in the synthetic counterpart represents the received power when applying a standard DFT codebook at the base station.} 
	\label{fig:dataset}
\end{figure*}

\textbf{CSI compression task:} The process of channel compression involves transforming a high-dimensional wireless channel, which can be hundreds or thousands of complex entries long, into a low dimensional representation, usually not more than a few tenths of real entries. Channel feedback, often bulky in its original form, is made more compact through this compression technique, thereby facilitating more efficient channel information transmission. A wireless channel can be a matrix $\mathbf{H}$ of $N_{BS}$ base station antennas by $N_{sub}$ subcarriers. Typically, channels used for compression are pre-processed through a transformation to the angle-delay domain by using Fourier transforms across each dimension of the matrix and flattening the resultant matrix. In this task, we consider $N_{BS} = 32$ and  $N_{sub} = 32$, and after transforming the channel to angle-delay, we trim the last 16 delay taps since they contain almost no information. The encoded dimension is taken as $N_{enc} = 32$, achieving a \(32\) times compression rate. As for the performance metric $\mathcal{P}$, it is common reconstruction tasks to use the normalized mean squared error (NMSE): 

\begin{equation}
	\text{NMSE}_{\text{dB}}(\textbf{H}, \hat{\textbf{H}}) = 10 \log_{10} \frac{\vert\vert \textbf{H} - \hat{\textbf{H}} \vert\vert^2_F}{\vert\vert \textbf{H}\vert\vert^2_F}.
\end{equation}

\textbf{Autoencoder model:} The model used to perform the channel reconstruction task is a autoencoder with the architecture of the model is presented in Figure \ref{fig:AE}. The figure illustrates a $32 \times 16$ complex channel matrix, which is expanded to \(32 \times 16 \times 2\) by separating the real and imaginary parts. The matrix then passes through several convolutional layers and a fully connected layer before reaching the \(32\)-dimensional latent space, marked in orange. The decoder then takes this latent space and, using a fully connected layer, reconstructs it back into a high-dimensional space. After reshaping, it passes through convolutional layers and a refinement network, which is repeated three times, before outputting the reconstructed channel, which is compared to the input using a minimum squared error (MSE) loss. Importantly, the model is trained in two ways: first, for performance evaluation, an autoencoder is trained for each area. When tested on the same area, the model achieves losses below $-20 \text{dB}$ NMSE, but higher losses in untrained zones. Second, for distance computation, a single special model with five refinement nets instead of three is trained using data from all datasets, achieving strong compression in \(32\) dimensions, averaging $-20 \text{dB}$ NMSE across areas. This model is used for dimensionality reduction, transforming the input space into the encoded space for distance computation. Later, we compare the results of distances applied to the raw space and latent spaces, showing the autoencoder provides both efficient encoding and strong dataset distancing performance.

\textbf{Dataset:} We use a raytraced dataset of the Arizona State University (ASU) campus generated using the DeepMIMO. The dataset consists of approximately $130000$ data points, with around $90000$ representing users with valid channels (outside buildings) over an area measuring $410$ meters by 320 meters, at a resolution of a user per meter. The raytracing simulation accounts for several paths that can be line-of-sight, reflections, diffuse scattering, and diffraction, making the simulation quite comprehensive and more realistic. The fidelity of the dataset is showcased in Figure \ref{fig:dataset}, which compares the real geographic data from Google Earth with a digital rendering.

 \begin{figure*} [t]
    \centerline{\includegraphics[width=2\columnwidth]{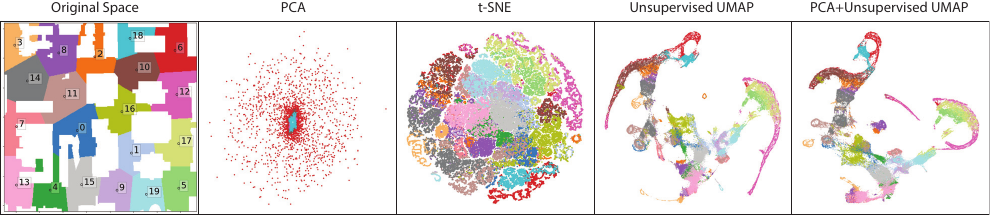}}
    \caption{Visualization of the latent spaces resultant from different dimensionality reduction methods applied to our dataset. From left to right: the original space where each dataset has been selected via proximity-based clustering, PCA, t-SNE, UMAP, and the combination of PCA (applied to reduce the dataset to $\sim100$ components) and UMAP.}
    \label{fig:unsup_comparison}
\end{figure*} 

\textbf{Distances in input space:} Table \ref{tab:raw_unsup} shows the results obtained by correlating the performances of the CSI compression task with the dataset distances obtained from several distance methods. We show three categories of distances (geometric, statistical and subspace) and the PAD. Since we previously defined clustering and dimensionality reduction methods as auxiliary tools for computing distances, in this table we omit the manifold methods, which rely on dimensionality reduction methods. Additionally, we include the clustered and centroid euclidean distances, which consider respectively, $3$ clusters and only the center of mass of $1$ cluster. We see these choices reflected in the computation times, since computing a mean is far faster than clustering all data. Overall, we note that Wasserstein and Energy distances (which are statistical IPMs) have superior performance, with $0.52$ and $0.55$ correlation with model performances. IPMs outperform $f$-divergences. Geometric distances suffer from curse of dimensionality, having low correlation. The same can be said about subspace distances, which perform poorly despite requiring computation resources. Additionally, poor performances of subspace distances can be attributed to the way principal angles are computed making it difficult to note differences in very similar subspaces. The best distance in the raw space is the PAD. The PAD learns how to categorize data, so it can work well when the dimensions are large and the data is easily separable. 

\begin{table}[t]
\label{tab:raw_unsup}
\caption{Compute time and correlation of distances computed in the raw space and model performances}
\centering
\begin{tabular}{|l|l|c|c|}
\hline
\textbf{Category} & \textbf{Distance Name}  & \textbf{Correlation} & \begin{tabular}[c]{@{}c@{}}\textbf{Compute}\\ \textbf{Time (s)}\end{tabular} \\ \hline
\multirow{4}{*}{Geometric} & Pairwise Euclidean      & 0.36                                      & 11                        \\ 
                           & Clustered Euclidean     & 0.37                                      & 166                       \\ 
                           & Centroid Euclidean      & 0.34                                      & 3                         \\ 
                           & Cosine                  & -0.07                                     & 5255                      \\ \hline
\multirow{8}{*}{Statistical} & Jensen-Shannon        & 0.14                                      & 80                        \\ 
                           & Hellinger               & 0.15                                      & 81                        \\ 
                           & Wasserstein             & \textbf{0.52}                             & 562                       \\ 
                           & Kolmogorov-Smirnov      & 0.47                                      & 381                       \\ 
                           & Total Variation         & 0.15                                      & 78                        \\ 
                           & MMD (linear)            & -0.08                                     & 79                        \\ 
                           & MMD (RBF)               & -0.06                                     & 135                       \\ 
                           & Energy                  & \textbf{0.55}                             & 735                       \\ \hline
\multirow{3}{*}{Subspace}  & Grassmann               & -0.11                                     & 15223                     \\ 
                           & Chordal                 & -0.10                                     & 14810                     \\ 
                           & Asimov                  & -0.03                                     & 15594                     \\ \hline
Other                      & PAD                     & \textbf{0.64}                             & 952                       \\ \hline
\end{tabular}
\end{table}

\textbf{Dimensionality reduction:} Space transformations such as dimensionality reduction methods can help distill information from the raw datasets into a space where distance computations are facilitated. We assess several of those methods in Figure \ref{fig:unsup_comparison}, by transforming $20$ datasets made from channels in positions geographically proximal to one another, into two-dimensional spaces, for convenient visualization. The goal is to obtain an intuitive understanding of these methods, and relate the $2$D representations with their characteristics. This figure shows that information encoded with PCA looks very similar across datasets, leading to no differentiation. t-SNE, on the other hand, is able to separate the datasets but it does not maintain the global structures. This is evident by noticing the datasets are mixed, and similar channels are not closer to each other. UMAP, on the other hand, maintains the local and global structures from the original data, while discarding almost all other information.

\textbf{Distances in latent spaces}: In the encoded spaces obtained through various dimensionality reduction techniques, we observe consistent trends in how distances reflect dataset similarities. Subspace distances remain small across these spaces, indicating that the datasets are closely aligned after encoding. However, most statistical distances—such as Jensen-Shannon, Hellinger, and Total Variation distances—do not perform as effectively. A possible reason for this underperformance is their limited range between 0 and 1, which can cap their sensitivity when dealing with disjoint distributions. In contrast, the Wasserstein distance does not have this limitation; it can scale to much larger values (e.g., in the thousands), making it more robust in distinguishing between datasets with significant distributional differences. 

\textbf{Comparing encoded spaces}: Our results demonstrate that in the PCA-encoded space, the PAD achieves the best performance among the evaluated metrics. This is likely because PCA, being a linear dimensionality reduction method, preserves the global structure of the data, allowing PAD to effectively differentiate between datasets. Although t-SNE reduces data to only two components, it often outperforms PCA on average, suggesting that t-SNE's emphasis on local data structure enhances its ability to organize data meaningfully. UMAP consistently surpasses t-SNE in all cases, with both Euclidean and Wasserstein distances performing particularly well in this space. The strong performance of Euclidean-based distances in UMAP implies that the encoded space is well-suited for distance computations, leveraging the intrinsic geometry established by UMAP manifold learning. Similarly, the Wasserstein distance excels due to its ability to capture distributional differences without the constraints of a limited range. Here, the autoencoder, trained across all zones, serves as a benchmark for maximum performance achievable through reconstruction-focused encoding. While it highlights the upper limits of performance, its practical utility may be limited compared to other methods in this context, since it requires training a separate models to compute distances. Moreover, as we show in Section \ref{sec:sup}, it does not perform nearly as well for supervised tasks. 

\textbf{Takeaway from unsupervised tasks}: The key takeaway is that choosing the right dimensionality reduction method can have a significant impact—sometimes even more than selecting the optimal distance metric to apply afterward. Nonetheless, Wasserstein and Euclidean distances are effective due to their robustness and simplicity, respectively. After evaluating dataset distances in an unsupervised task, we present results for two supervised tasks. This is particularly relevant because most machine learning tasks in wireless communications are supervised, yet many distance metrics do not treat labels as distinct elements of the dataset and process them similarly to inputs. To address this, we propose a new distance metric that incorporates labels, which may outperform methods that ignore label information.

\begin{table}[t]
\setlength{\tabcolsep}{0.4em} 
\centering
\caption{Correlation between model performances and distances computed in a reduced space obtained form retaining either 2 or 32 components of different dimensionality reduction schemes, namely PCA, TSNE, UMAP and an autoencoder.}
\begin{tabular}{|c|l|cccc|}
\hline
\multirow{2}{*}{\textbf{Category}}          & \multirow{2}{*}{\textbf{Distance}} & \multicolumn{4}{c|}{\textbf{Dimensionality Reduction}} \\ 
                                            &                                        & \textbf{PCA 32} & \textbf{TSNE 2} & \textbf{UMAP 2} &  \textbf{AE 32} \\ \hline
\multirow{4}{*}{\textbf{Geometric}}         & Pairwise Euclidean                     & 0.37            & 0.58            & \textbf{0.83}   & \textbf{0.87}    \\ 
                                            & Clustered Euclidean                    & 0.41            & 0.59            & \textbf{0.84}   & \textbf{0.91}    \\ 
                                            & Centroid Euclidean                     & 0.35            & 0.59            & \textbf{0.86}   & \textbf{0.93}    \\ 
                                            & Cosine                                 & 0.30            & 0.41            & 0.42            & \textbf{0.94}    \\ \hline
\multirow{9}{*}{\textbf{Statistical}}       & KL Divergence                          & 0.32            & 0.62            & 0.52            & \textbf{0.85}    \\ 
                                            & Jensen-Shannon                         & -0.08           & 0.15            & 0.12            & 0.07             \\ 
                                            & Hellinger                              & -0.08           & 0.17            & 0.13            & 0.07             \\ 
                                            & Wasserstein                            & 0.47            & 0.68            & \textbf{0.85}   & \textbf{0.92}    \\ 
                                            & Kolmogorov-Smirnov                     & 0.57            & 0.32            & 0.46            & 0.22             \\ 
                                            & Total Variation                        & -0.06           & 0.13            & 0.10            & 0.04             \\ 
                                            & MMD (Linear)                           & -0.17           & 0.10            & 0.04            & 0.04             \\ 
                                            & MMD (RBF)                              & -0.13           & 0.06            & 0.04            & -0.02            \\ 
                                            & Energy                                 & 0.56            & 0.42            & 0.60            & 0.25             \\ \hline
\multirow{3}{*}{\textbf{Subspace}}          & Grassmann                              & -0.14           & 0.02            & -0.22           & -0.05            \\ 
                                            & Chordal                                & -0.14           & 0.02            & -0.22           & -0.05            \\ 
                                            & Asimov                                 & -0.12           & 0.03            & -0.23           & -0.06            \\ \hline
\textbf{Other}                              & PAD                                    & \textbf{0.75}   & 0.68            & 0.71            & 0.66             \\ \hline
\end{tabular}
\end{table}

\begin{figure*} [t]
    \centerline{\includegraphics[width=2\columnwidth]{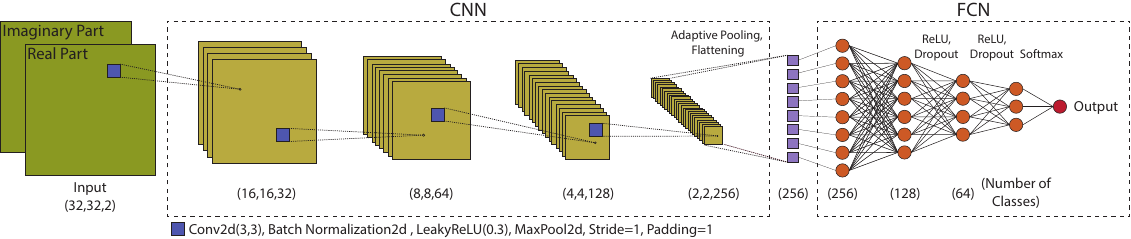}}
    \caption{The CNN model used for supervised tasks}
    \label{fig:CNN}
\end{figure*}

\section{Label-Aware Dataset Distance}

In this section, we introduce a novel metric called the \textbf{label-aware dataset distance}, which incorporates label information to enhance the computation of distances between datasets with labels. Traditionally, distances have been calculated solely based on input features, or by grouping inputs with labels, which leads to a loss of the relationships between inputs and outputs. However, by intentionally utilizing label information, we can achieve a more informed measure of dataset similarity, especially when the label distributions differ between datasets.

\textbf{Applicability}: The label-aware method can be applied in both encoded spaces (such as those generated by UMAP) and raw feature spaces. The key component of this method is the computation of a \textbf{penalty term} to address unbalanced label distributions. A pair of datasets is considered unbalanced when one dataset contains labels that the other does not. For each label \( l \) in the union of labels \( L = L_1 \cup L_2 \), where \( L_i \) is the set of labels in dataset \( D_i \), we define a penalty term  \( P_l \) and a label-specific distance \( d_l \).

\textbf{Penalty Term \( P_l \):} The penalty term for label \( l \) is set as the maximum distance between any two points with label \( l \) across all datasets under consideration. Let there be \( S \) datasets in total, and let \( D_s^l \) denote the subset of dataset \( D_s \) containing only samples with label \( l \), where \( s \in \{1, 2, \dots, S\} \). The penalty term is defined as:

\[
P_l = \max_{\substack{\mathbf{x} \in D_p^l, \ \mathbf{y} \in D_q^l \\ p,q \in \{1,2,\dots,S\}}} d(\mathbf{x}, \mathbf{y}),
\]

where \( d(\mathbf{x}, \mathbf{y}) \) is the chosen distance metric (e.g., Euclidean distance in the encoded space). This penalty ensures consistency across the group of datasets being considered and accounts for label-specific variations, as some labels might have inherently larger distances.

\textbf{Label-Specific Distance \( d_l \):} For labels present in both datasets \( D_1 \) and \( D_2 \), we compute the distance using only the samples with label \( l \):

\[
d_l = d\left(D_1^l, D_2^l\right),
\]

where \( d\left(D_1^l, D_2^l\right) \) represents the distance computation method (e.g., the pairwise euclidean distance) applied to the subsets corresponding to label \( l \). This effectively formulates the problem as an unsupervised distance computation within each label class. For labels that are present in one dataset but not the other, we assign \( d_l = P_l / 2\) to penalize the imbalance

\[
d_l = 
\begin{cases}
d\left(D_1^l, D_2^l\right), & \text{if } l \in L_1 \cap L_2, \\
P_l / 2, & \text{if } l \in L_1 \Delta L_2,
\end{cases}
\]
where \( \Delta \) denotes the symmetric difference between sets. 

\textbf{Aggregating label-specific distances:} After computing the distances for each label, we aggregate them to obtain the overall label-aware dataset distance \( d_{\text{label-aware}} \):

\[
d_{\text{label-aware}} = \frac{1}{|L|} \sum_{l \in L} d_l,
\]
where \( |L| \) is the total number of unique labels in the union of \( L_1 \) and \( L_2 \). Labels absent in both datasets are not considered in the computation, ensuring that only relevant labels influence the distance.

By incorporating the penalty term for labels absent in one of the datasets, the label-aware distance effectively penalizes the unbalanced label distributions, reflecting the potential challenges in transferring models between such datasets. A straightforward example would be a model trained only in LoS data struggling to predict NLoS labels. This method allows for a more nuanced comparison between datasets, accounting for both the presence and distribution of labels, and can be integrated with existing distance computation methods in either encoded or raw feature spaces. In the next section we show how this method compares with other methods applied to supervised tasks.

\section{Supervised case: Beam Prediction} \label{sec:sup}

\begin{figure*} [t]
    \centerline{\includegraphics[width=1.95\columnwidth]{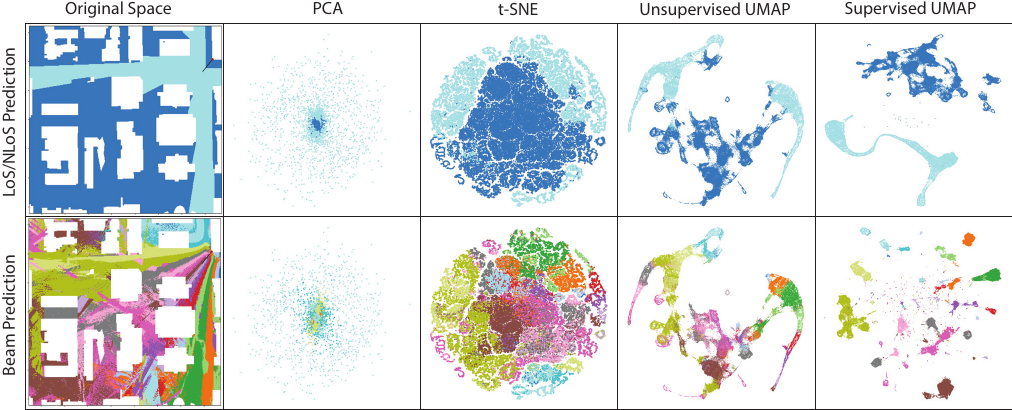}}
    \caption{Comparison of latent spaces (PCA, t-SNE, UMAP) with original space  in XY/cartesian coordinates. Colors mean different labels (LoS status or beam index). Figure shows that UMAP provides a clearer separation with less overlap while maintaining a more truthful global structure.}
    \label{fig:sup_comparison}
\end{figure*} 

In this section, we apply dataset distance measures to a supervised machine learning task, specifically beam prediction. The dataset used for these supervised tasks is the same as the one employed in the unsupervised case, augmented with the appropriate labels for each task. By applying dataset distance measures to these supervised learning problems, we aim to assess how well methods that worked well in unsupervised tasks perform in supervised. Moreover, we analyze the benefit of separately considering the label information via our proposed label-aware methods. In summary, the variables defined in Section \ref{sec:problem} take the following values:

\begin{itemize}
	\item $\mathcal{T}$: Task: Optimum beam prediction
	\item $\mathcal{P}$: Accuracy (\%)
	\item $\mathcal{M}$: Deep convolutional net, architecture in Figure \ref{fig:CNN}
	\item $\boldsymbol{\mathcal{D}}$: $20$ areas from ASU dataset and respective task labels
\end{itemize}

\textbf{Supervised Tasks}: 
The beam prediction task 
($\mathcal{T}$) requires applying a set of beamformers to the channels and identifying which beamformer yields the highest received power at the user end. In this scenario, we utilize an antenna array with $32$ elements and a corresponding codebook of $32$ beams; each beam serves as a distinct label for classification. For both tasks, we employ the same performance metric, the top-$1$ accuracy, which measures the percentage of instances where the model correctly predicts the exact label. Mathematically, we can define the top-$1$ accuracy \( A_{\text{top-1}} \) is defined as:

\[
A_{\text{top-1}} = \frac{1}{N} \sum_{i=1}^{N} \delta(y_i, \hat{y}_i)
\]

where \( N \) is the total number of samples, \( y_i \) is the true label for the \( i \)-th sample and \( \hat{y}_i \) is the predicted label for the \( i \)-th sample. Note, here \( \delta(y_i, \hat{y}_i) \) represents the indicator function, such that

\[
\delta(y_i, \hat{y}_i) = 
\begin{cases}
1, & \text{if } y_i = \hat{y}_i \\
0, & \text{if } y_i \neq \hat{y}_i
\end{cases}
\]

The leftmost graphs of Figure \ref{fig:sup_comparison} shows how each label looks in the ASU dataset and how each dimensionality reduction method changes the space of labels. The datasets are divided the exact same way as before as well, so a given area/dataset can have multiple different labels, depending on the location of the dataset and specific task.

\textbf{Convolutional model}: To accomplish these tasks, we use a convolutional neural network (CNN), as depicted in Figure \ref{fig:CNN}. The network architecture comprises multiple layers of $2$D convolutions, batch normalization, leaky ReLU activations, and max pooling. These operations are applied to both the real and imaginary components of the input data. The final layers leverage fully connected networks that process the extracted features and output softmax-normalized class probabilities corresponding to the number of labels in each task.

\begin{table}[t]
\centering
\caption{Correlation of raw distances with model performance in supervised tasks.}
\begin{tabular}{|l|l|cc|}
\hline
\multirow{2}{*}{\textbf{Category}} & \multirow{2}{*}{\textbf{Distance Name}} & \multicolumn{2}{c|}{\textbf{Correlation}} \\ 
                                            &                                        & \textbf{LOS ID}   & \textbf{Beam Pred.}    \\ \hline
\multirow{4}{*}{Geometric}                  & Pairwise Euclidean                     & 0.43              & 0.07                  \\ 
                                            & Clustered Euclidean                    & 0.42              & 0.12                  \\ 
                                            & Centroid Euclidean                     & 0.32              & 0.06                  \\ 
                                            & Cosine                                 & -0.05             & -0.05                 \\ \hline
\multirow{8}{*}{Statistical}                & Jensen-Shannon                         & 0.08              & 0.14                  \\ 
                                            & Hellinger                              & 0.09              & 0.15                  \\ 
                                            & Total Variation                        & 0.07              & 0.14                  \\ 
                                            & Wasserstein                            & \textbf{0.55}              & 0.23                  \\ 
                                            & Kolmogorov-Smirnov                     & 0.50              & 0.42                  \\ 
                                            & Energy                                 & \textbf{0.63}              & 0.33                  \\ 
                                            & MMD (Linear)                           & -0.10             & -0.01                 \\ 
                                            & MMD (RBF)                              & -0.12             & -0.01                 \\ \hline
\multirow{3}{*}{Subspace}                   & Grassmann                              & -0.09             & -0.08                 \\ 
                                            & Chordal                                & -0.08             & -0.07                 \\ 
                                            & Asimov                                 & 0.12              & -0.06                  \\ \hline
Other                                       & PAD                                    & 0.45              & \textbf{0.61}         \\ \hline
\multirow{2}{*}{\textbf{Proposed}}          & Label-aware Euclidean                  & 0.18              & \textbf{0.75}         \\
                                            & Label-aware Wasserstein                & \textbf{0.53}              & 0.11                  \\ \hline
\end{tabular}
\end{table}

\textbf{Distances in input space}: In the input space, our results indicate that certain distances, such as the Proxy-A Distance and IPMs like the Wasserstein, Kolmogorov-Smirnov, and Energy distances, are effective for measuring dataset similarities. The proposed label-aware Euclidean distance performs well, particularly in the beam prediction task, where label imbalances are less. In contrast, the label-aware Wasserstein distance does not perform as effectively, which can be attributed to the high dimensionality of the raw input space. Additionally, the line-of-sight (LoS) identification task poses significant challenges because many datasets contain only a single label. For this particular task, we show the results only for the datasets with at least a sample of each label, since models cannot generalize to datasets with unseen labels. Label-aware methods alleviate this issue by introducing penalty terms based on label imbalances. However, they are still affected by the curse of dimensionality and since the penalty terms need to be estimated, label-aware methods are tied to the task and dataset. 

\textbf{Distances in latent spaces}: When examining the results from dimensionality reduction methods, we observe that techniques encoding global relationships in the data, such as UMAP and autoencoders, yield better performance on average. PCA, being a linear method, does not capture these global structures and therefore works well primarily with linear distances like Euclidean distance and PAD. In scenarios where a linear dimensionality reduction method is preferred, the label-aware Euclidean distance outperforms other metrics. We also find that label-aware distances tend to surpass other methods or provide similar performances. Interestingly, supervised UMAP, which uses labels to enhance data separation, performs slightly better than UMAP in cases with few labels and worse in cases with many labels — this may be attributed to an excessive emphasis on label information leading to an overseparation of data points beyond their natural distinctions in the input space. Overall, we observe that label-aware approaches boost performance compared to both Euclidean and Wasserstein distances, and correlation scores in encoded spaces are higher compared to raw spaces. 

\textbf{Conclusion from supervised tasks}: Results suggest the effectiveness of label-aware distances. Our previous methods utilizing Wasserstein or Euclidean distances in UMAP-encoded spaces remain promising for unsupervised cases. However, label-aware distances present greater potential gains when the label information is utilized effectively. Despite the good performance in our experiments, we acknowledge that improvements could be made to the algorithms leveraging label information. In particular, the penalties are too dependent on the problem, and not yet fully generalizable. Nonetheless, we recommend the use of label-aware euclidean for tasks with many labels and Wasserstein for scenarios with fewer labels, and opting for euclidean when in doubt since it still performs well in most cases with distinctively low complexity.

\section{Conclusion} \label{sec:conclusion}

In this work, we introduced a comprehensive framework for dataset distancing in wireless communications, establishing one of the first direct correlations between dataset distances and model accuracy. By leveraging latent spaces derived from dimensionality reduction techniques like PCA, t-SNE, UMAP, and autoencoders, we effectively captured intrinsic data structures, enabling robust distance computations in both unsupervised and supervised tasks. We further proposed a novel label-aware dataset distance metric that incorporates label information into the distance computation process, providing a more robust measure of similarity. Our extensive evaluations demonstrated that our proposed distances outperform traditional methods in raw spaces and across various dimensionality reduction techniques, particularly in tasks like line-of-sight identification and beam prediction. This foundational step towards understanding and quantifying dataset similarities can enhance data selection, reduce unnecessary model retraining, and facilitate more efficient deployment of machine learning models in wireless systems. Future work will build upon these findings, expanding our methods to a broader range of wireless tasks, applying them to real-world datasets, and integrating them with advanced models to further advance the field.

\begin{table}[t]
\setlength{\tabcolsep}{0.26em} 
\centering
\caption{Correlation of distances in dimensionality reduced spaces with model performance in supervised tasks.}
\begin{tabular}{|c|l|ccccc|}
\hline
\multirow{3}{*}{\textbf{Task}}    & \multirow{3}{*}{\hspace{.6cm}\textbf{Distance}}      & \multicolumn{5}{c|}{\textbf{Dimensionality Reduction Method}}                               \\ 

                                   &                                         & \multirow{2}{*}{\textbf{PCA 32}} & \multirow{2}{*}{\textbf{TSNE 2}} & \multirow{2}{*}{\textbf{UMAP 2}} & \multirow{2}{*}{\textbf{AE 32}} & \multirow{2}{*}{\textbf{\makecell{SUP\\UMAP 2}}} \\

                                  & & & & & & \\               \hline                   
\multirow{8}{*}{\rotatebox{90}{\textbf{LoS ID}}}  & Pairwise Euclidean                     & 0.43            & 0.38            & 0.46            & 0.41            & \textbf{0.73}                 \\ 
                                  & Clustered Euclidean                  & 0.40            & 0.45            & 0.51            & 0.49            & \textbf{0.63}                 \\ 
                                  & Centroid Euclidean                 & 0.35            & 0.32            & 0.47            & 0.47            & \textbf{0.68}                 \\ 
                                  & KL Divergence                           & 0.23            & 0.35            & 0.17            & 0.51            & 0.21                 \\ 
                                  & Wasserstein                             & 0.50            & 0.43            & 0.45            & 0.47            & \textbf{0.71}                 \\ 
                                  & PAD                                     & 0.44            & 0.38            & 0.37            & 0.34            & 0.36                 \\ 
                                  & Label-aware Euclid.                    & 0.10            & 0.22            & \textbf{0.78}            & 0.40            & \textbf{0.80}                 \\ 
                                  & Label-aware Wasser.                 & 0.28            & 0.25                & \textbf{0.75}            & 0.45            & \textbf{0.81}                 \\ \hline
\multirow{8}{*}{\rotatebox{90}{\textbf{Beam Prediction}}} & Pairwise Euclidean             & 0.07            & 0.62            & 0.73            & \textbf{0.76}   & 0.63                 \\ 
                                  & Clustered Euclidean                  & 0.12            & 0.56            & 0.71            & \textbf{0.75}   & 0.56                 \\ 
                                  & Centroid Euclidean                 & 0.05            & 0.70            & \textbf{0.79}   & \textbf{0.79}   & 0.72                 \\ 
                                  & KL Divergence                           & 0.25            & 0.60            & 0.50            & 0.72            & 0.74                 \\ 
                                  & Wasserstein                             & 0.15            & 0.72            & \textbf{0.76}   & \textbf{0.79}   & 0.72                 \\ 
                                  & PAD                                     & \textbf{0.73}            & 0.66            & 0.67            & 0.63            & 0.70                 \\ 
                                  & Label-aware Euclid.                   & \textbf{0.74}            & \textbf{0.78}   & \textbf{0.79}   & \textbf{0.81}   & \textbf{0.81}        \\ 
                                  & Label-aware Wasser.                 & 0.19            & \textbf{0.77}   & \textbf{0.80}   & \textbf{0.79}   & \textbf{0.81}        \\ \hline
\end{tabular}
\end{table}

\section{Future Work}

Future work should address two limitations: (i) many distances require retraining an embedding model, which hinders online use, and (ii) UMAP-style mappings often require access to the full dataset and can be costly to scale. A natural direction is to learn \textbf{scalable, reusable embedding spaces} via \textbf{parametric neural mappings} trained with mini-batch objectives, enabling streaming/continual updates. In supervised settings, label-aware encoders can shape the latent space to better reflect transferability. Large Wireless Models (LWM) \cite{alikhani2025largewirelessmodellwm} offer a path to reduce per-task retraining by providing universal embeddings: datasets can be compared by embedding new samples once and computing distances in a shared space, with only lightweight calibration when needed. Additionally, the methods evaluated here for a limited set of tasks can be extended and improved to a broader range of wireless tasks, including those with mixed inputs—such as combining signal-to-noise ratio (SNR) with channel data, or incorporating angles and times of arrival. Applying our techniques to real-world data is also a crucial step, particularly towards achieving ray-traced site-specific dataset generation that closely mirrors actual deployment scenarios. Another promising avenue is the creation of distance-informed loss functions for generative models \cite{generativeAIEra}, which can enhance the quality and relevance of generated data. Furthermore, our methods can be extended to tasks involving discrete or categorical input variables. Ultimately, implementing end-to-end applications like data generation, dataset ranking, dataset compression, and outlier or dataset shift detection should be a significant focus. With these advancements, we envision a more intentional use of datasets leading to better data-driven approaches. 

\balance
\bibliographystyle{IEEEtran}
\bibliography{biblio}

\end{document}